\documentclass{article}
\usepackage{ijcai23}

\usepackage{times}
\usepackage{soul}
\usepackage{url}
\usepackage[hidelinks]{hyperref}
\usepackage[utf8]{inputenc}
\usepackage[small]{caption}
\usepackage{graphicx}
\usepackage{amsmath}
\usepackage{amsthm}
\usepackage{booktabs}
\usepackage{algorithm}
\usepackage{algorithmic}
\usepackage[switch]{lineno}
\usepackage{bm}
\usepackage{natbib}
\usepackage{amsfonts}
\usepackage{caption}
\usepackage{enumitem}
\usepackage{multirow}
\usepackage{makecell}
\usepackage{subfigure}
\usepackage{xspace}

\makeatletter
\DeclareRobustCommand\onedot{\futurelet\@let@token\@onedot}
\def\@onedot{\ifx\@let@token.\else.\null\fi\xspace}
 
\def\eg{\emph{e.g}\onedot} 
\def\ie{\emph{i.e}\onedot}

\makeatother

\title{A Survey on Spectral Graph Neural Networks}

\author{
Deyu Bo$^1$\and
Xiao Wang$^1$\and
Yang Liu$^1$\and
Yuan Fang$^2$\and
Yawen Li$^1$\And
Chuan Shi$^1$\thanks{Corresponding author}
\affiliations
$^1$Beijing University of Posts and Telecommunications, Beijing, China\\
$^2$Singapore Management University, Singapore\\
\emails
\{bodeyu, xiaowang, liuyangjanet, shichuan\}@bupt.edu.cn, yfang@smu.edu.sg, lywbupt@126.com
}

\begin{document}

\maketitle

\begin{abstract}
Graph neural networks (GNNs) have attracted considerable attention from the research community.
It is well established that GNNs are usually roughly divided into spatial and spectral methods.
Despite that spectral GNNs play an important role in both graph signal processing and graph representation learning, existing studies are biased toward spatial approaches, and there is no comprehensive review on spectral GNNs so far.
In this paper, we summarize the recent development of spectral GNNs, including model, theory, and application.
Specifically, we first discuss the connection between spatial GNNs and spectral GNNs, which shows that spectral GNNs can capture global information and have better expressiveness and interpretability.
Next, we categorize existing spectral GNNs according to the spectrum information they use, \ie, eigenvalues or eigenvectors.
In addition, we review major theoretical results and applications of spectral GNNs, followed by a quantitative experiment to benchmark some popular spectral GNNs. Finally, we conclude the paper with some future directions.
\end{abstract}

\section{Introduction}

Graph neural networks (GNNs) have shown strong performance in graph-related tasks, including node classification \citep{GCN, GAT}, link prediction \citep{SEAL}, and graph classification \citep{GIN}.
In general, GNNs can be divided into spatial and spectral methods according to different definitions of graph convolution.
Spatial methods, also known as message-passing neural networks (MPNNs) \citep{MPNN}, perform graph convolutional in the vertical domain by aggregating the neighborhood information along graph structures.
In contrast, spectral methods use spectral graph theory \citep{SGT} to transform the convolutions into products of the frequency domain.
Over the past few years, research on GNNs has focused on spatial methods because of their flexibility and scalability, \citep{GraphSAGE}.
On the contrary, research on spectral methods is somewhat under-explored.

Spectral GNNs have related but different views from spatial methods and play an important role in graph representation learning.
First, \textbf{spatial and spectral GNNs capture different information}.
Spatial GNNs aggregate node features layer by layer. Nodes can only capture information within a fixed distance, emphasizing local information.
In contrast, spectral GNNs transform all node features into weighted sums of different eigenvectors via graph Fourier transform, which naturally captures global information.
Second, \textbf{spatial and spectral GNNs use different design principles}.
Spatial GNNs involve node-based updating, where the gradients flow between the connected nodes only and therefore have lower complexity, but their expressive power is bounded from above by the 1-Weisfeiler-Lehman (WL) test \citep{GIN}.
Spectral GNNs involve feature-based updating, where each feature is filtered by the fully-connected eigenspaces, and messages are passed among all nodes simultaneously.
Although more complex, feature-based updating breaks the localized property and is thus more powerful than node-based updating \citep{GNNML}.
Third, \textbf{spatial and spectral GNNs have different interpretability}.
Spatial GNNs require a post-hoc interpretation strategy, which aims to find the most important graph structures related to a prediction, \eg, nodes, edges, or subgraphs \citep{GNNExplainer}.
Spectral GNNs are interpretable models in which the learned graph filters can directly state the most important frequency information associated with the labels, \eg, low-, medium-, and high-frequencies.


The above three points show that spectral GNNs are important in designing and explaining GNNs.
However, little effort has been made to summarize the development of spectral methods.
On the one hand, existing surveys of GNNs \citep{survey1, survey2} focus on spatial methods.
On the other hand, spectral GNNs are close to graph signal processing (GSP). \citet{GSP2} and \citet{GSP1} are two surveys on traditional GSP  that formally define some fundamental concepts, such as filtering and sampling, but they do not explain the connection between spectral filtering and graph representation learning in detail.
Besides, while both eigenvalues and eigenvectors play a role in the spectral domain, the latter has been largely ignored in previous literature.

In this paper, we provide a comprehensive and systematic review of spectral GNNs.
Specifically, we first analyze the differences between spatial and spectral GNNs and introduce the unique challenges of designing spectral methods.
Then, we differentiate existing methods into two categories: eigenvalue-based and eigenvector-based, which correspond to the design of filters and basis functions in signal processing.
Within each category, there are three subcategories, and we further analyze their strengths and weaknesses in terms of effectiveness and efficiency.
Next, we introduce major theoretical results and important applications of spectral GNNs. We also make a quantitative experiment to test the performance and complexity of different graph filters on both homophilic and heterophilic benchmarks. Finally, we discuss a few promising future directions in spectral GNNs.

\section{Challenges and Concepts}

Spectral GNNs utilize neural networks to learn representations from spectral information. The design of spectral GNNs is not trivial, which poses the following three challenges:
\begin{itemize}
    \item \textbf{Diversity of spectral information.}
    Spectral GNNs aim to learn node or graph representations from spectral information, which is informative but hard to explore.
    For example, eigenvalues and eigenvectors indicate the global and local structures \citep{GPS}, and the Fourier coefficients reflect the energy of signals on different harmonics.
    These spectral features with different meanings and tensor sizes pose the first challenge.
    \item \textbf{Complexity of graph data.}
    There are various types of graph data in real-world applications, such as dynamic graphs, heterogeneous graphs, hyper-graphs, etc.
    Most of them cannot be represented by a basic adjacency matrix.
    Therefore, how to extend the idea of spectral convolution to graph data other than undirected graphs is indispensable for spectral GNNs.
    \item \textbf{Scalability of methodology.}
    The complexity of the eigenvalue decomposition is the cube of the number of nodes.
    Besides, explicitly constructing the fully-connected eigenspaces also brings a quadratic time and space complexity.
    Both operations are important but expensive for spectral GNNs.
    How to make spectral GNNs scalable to large-scale graphs is a great challenge for spectral GNNs.
\end{itemize}
The reviews of spectral GNNs are shown in Section 3, where each approach solves at least one of these three challenges.
Before that, we first introduce some important concepts to give a general understanding of spectral GNNs.

\begin{figure}[t]
    \centering
    \includegraphics[width=0.9\linewidth]{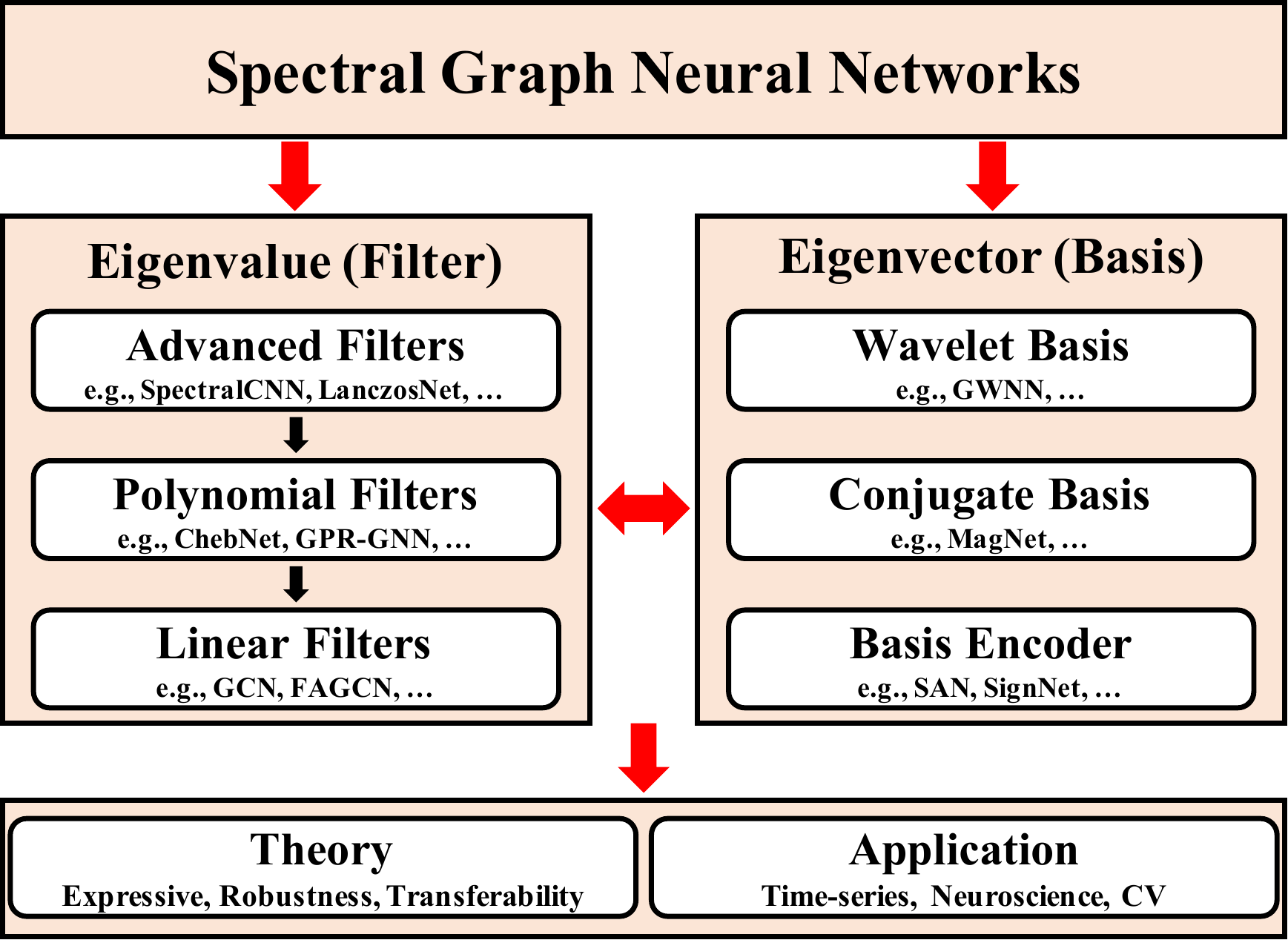}
    \caption{Framework of this survey.}
    \label{fig: framework}
\end{figure}

Assume that $\mathcal{G}=(\mathcal{V}, \mathcal{E})$ is a graph, where $\mathcal{V}$ is the node set with $|\mathcal{V}|=N$ and $\mathcal{E}$ is the edge set.
Let $\mathbf{A} \in \{0, 1\}^{N \times N}$ be the adjacency matrix of $\mathcal{G}$ and the normalized graph Laplacian matrix is defined as $\mathbf{L}=\mathbf{I}_{N}-\mathbf{D}^{-1/2}\mathbf{A}\mathbf{D}^{-1/2}$, where $\mathbf{D}$ is a diagonal degree matrix with $D_{i,i}=\sum_{j}A_{i,j}$ and $\mathbf{I}_{N}$ denotes the identity matrix.

\paragraph{Eigenvalue Decomposition (EVD).}
The graph Laplacian matrix can be decomposed as $\mathbf{L} = \mathbf{U} \bm{\Lambda} \mathbf{U}^{\top}$, where $\bm{\Lambda}=\text{diag}\{\lambda_{1},\cdots,\lambda_{N}\}$ is a diagonal matrix of eigenvalues and $\mathbf{U}=[\mathbf{u}_{1}, \cdots, \mathbf{u}_{N}]$ is the corresponding eigenvectors.
If $\mathbf{L}$ is symmetric, the eigenvalues are real and satisfy $0 \leq \lambda_{1} \leq \cdots \leq \lambda_{N} \leq 2$. Otherwise, both $\bm{\Lambda}$ and $\mathbf{U}$ will be complex.
If no special emphasis, the graphs are undirected.

\paragraph{Graph Fourier Transform (GFT).}
Discrete Fourier transform (DFT) aims to decompose an input signal or function into a weighted sum of the orthogonal bases.
Through SVD, we have $\mathbf{L} \mathbf{u}_{i} = \lambda_{i} \mathbf{u}_{i}$ and $\lambda_{i} = \mathbf{u}_{i}^{\top} \mathbf{L} \mathbf{u}_{i}$.
Note that the eigenvalues reflect the frequency/smoothness of the corresponding eigenvectors.
Since the eigenvectors are normalized and orthogonal, \ie $||\mathbf{u}_{i}||=1$ and $\mathbf{u}_{i}^{\top}\mathbf{u}_{i}=\delta_{ij}$, the eigenvectors can be used as the orthogonal bases.
Given a graph signal $\mathbf{x} \in \mathbb{R}^{N \times 1}$, its GFT and inverse GFT can be defined as:
\begin{equation}
    \hat{\mathbf{x}}=\mathbf{U}^{\top}\mathbf{x}, \
    \mathbf{x}=\mathbf{U}\hat{\mathbf{x}},
\end{equation}
where $\hat{\mathbf{x}}$ is the Fourier coefficients of $\mathbf{x}$.

Based on the eigenvalues and eigenvectors of graph Laplacian, we can define the convolutional in the Fourier domain.

\paragraph{Spectral Graph Convolution.}
The convolution between a filter $f$ and a signal $\mathbf{x}$ in the spatial domain is equal to their production in the spectral domain.
Through GFT, the spectral graph convolutional is defined as
\begin{equation}
\label{convolution}
f *_{G} \mathbf{x} = \mathbf{U} \left( \left( \mathbf{U}^{\top}f \right) \odot \left( \mathbf{U}^{\top}\mathbf{x} \right) \right),
\end{equation}
where $f *_{G} x$ is the convolution and $\odot$ is the production.
Generally, spectral GNNs will parameterize $\mathbf{U}^{\top}f$ with a learnable diagonal matrix $\mathbf{G}=\text{diag}\{g_{1}, \cdots, g_{N}\}$ as the spectral filter that can be optimized by gradient descent.
We can rewrite the spectral graph convolutional as:
\begin{equation}
    f *_{G} \mathbf{x} = \mathbf{U} \mathbf{G} \mathbf{U}^{\top} \mathbf{x} = (g_{1}\mathbf{u}_{1}\mathbf{u}_{1}^{\top} + \cdots +g_{N}\mathbf{u}_{N}\mathbf{u}_{N}^{\top})\mathbf{x},
\label{eq: ugux}
\end{equation}
where $\mathbf{u}_{i}\mathbf{u}_{i}^{\top} \in \mathbb{R}^{N \times N}$ is the $i$-th fully-connected eigenspace.

\section{Spectral Graph Neural Networks}

In this section, we classify existing spectral GNNs into two categories: eigenvalue-based and eigenvector-based, according to the spectral information they used.
Within each category, we will introduce three subcategory methods and explain their advantages and disadvantages as well as the connection in detail.
Additionally, we present advances in theory and application, and benchmark existing methods on six datasets.
The overall framework can be seen in Figure \ref{fig: framework}.

\subsection{Eigenvalue-based Spectral GNNs}

Eigenvalue-based spectral GNNs correspond to the design of filters in signal processing.
Approaches belonging to this category aim to learn a good filter that captures the most important frequency information of input graph signals.
In this section, we introduce three types of graph filters: advanced filters, polynomial filters, and linear filters, from which we can observe how the spectral GNNs evolve to reduce complexity and scale to large graphs, along with the declining of expressive power.

\subsubsection{Advanced Filters}
We refer to the spectral GNNs that explicitly take eigenvalues as the input of neural networks as advanced filters.
Since the decomposition of graph Laplacian is time-consuming, advanced filters usually choose to truncate the graph spectrum, \ie, using the smallest-$k$ eigenvalues and eigenvectors, to reduce the complexity.
Although computationally intensive, such methods can capture the geometric information hidden in the eigenvalues.
For example, the second-smallest eigenvalue indicates the algebraic connectivity of a graph.

SpectralCNN \citep{SpectralCNN} is the first advanced graph filter, which treats the graph filter as the parameters of neural networks and directly optimizes it by gradient descent:
\begin{equation}
    \mathbf{h}_{j}^{(l+1)} = \sigma(\mathbf{U} \sum_{i=1}^{d_{in}}\mathbf{G}_{ij}^{(l)} \mathbf{U}^{\top}\mathbf{h}_{i}^{(l)}) \quad (j=1 \cdots d_{out}),
\end{equation}
where $\sigma$ is the activation function and $\mathbf{h}_{i}^{(l)} \in \mathbb{R}^{N \times 1}$ indicates the $i$-th dimension of the representations in $l$-th layer.
SpectralCNN generalizes the design of convolutional neural networks (CNNs) to graphs and assigns each feature an independent graph filter, \ie $\mathbf{G}_{ij}^{(l-1)}$.
Therefore, the total parameters of one convolution layer are $\mathcal{O}(k^{2} d_{in} d_{out})$.
More parameters enhance the fitting ability of SpectralCNN but also bring high time and space complexity.
Additionally, the non-parametric graph filters make the model brittle. When there are multiple eigenvalues, SpectralCNN cannot handle the ambiguity of eigenvectors \citep{PEG}.

LanczosNet \citep{LanczosNet} uses spectral convolution to define a multi-scale graph representation learning framework.
Spatial GNNs construct multi-scale representations by stacking multiple message-passing layers, \ie, calculating $\mathbf{A}, \mathbf{A}^{2}, \cdots, \mathbf{A}^{K}$ recurrently, which is infeasible for large graphs.
LanczosNet leverages the fact $\mathbf{A}^{K} = \mathbf{U} \bm{\Lambda}^{K} \mathbf{U}^{\top}$ to reduce complexity and 
capture the long-range information in the spectral domain, which is more efficient. The long-range spectral filter is defined as:
\begin{equation}
    L_{i}(\mathcal{I}) = \sum_{t=1}^{k}f_{i}(\lambda_{t}^{\mathcal{I}_{1}}, \lambda_{t}^{\mathcal{I}_{2}}, \cdots, \lambda_{t}^{\mathcal{I}_{|\mathcal{I}|}}) \mathbf{u}_{t}\mathbf{u}_{t}^{\top},
\end{equation}
where $f_{i}$ is an element-wise function, such as multilayer perceptron (MLP) and $\mathcal{I}=\{10, 20, \cdots, 50\}$.
For short-range information, LanczosNet directly uses spatial convolution. The overall framework is defined as:
\begin{equation}
    \mathbf{h}^{(l+1)} = [\mathbf{A}\mathbf{h}^{(l)} \cdots \mathbf{A}^{K}\mathbf{h}^{(l)}, L_{1}(\mathcal{I})\mathbf{h}^{(l)} \cdots L_{T}(\mathcal{I})\mathbf{h}^{(l)}]\mathbf{W}.
\end{equation}
It is worth noting that LanczosNet takes the eigenvalues as input and outputs the long-range graph filter.
By leveraging the correspondence between eigenvalues and eigenvectors, LanczosNet learns invariant representations to the permutation of nodes and eigenvectors.
The design of parameterizing eigenvalues as graph filters is first proposed by \cite{ChebNet} and continued in subsequent work.

Specformer \citep{Specformer} is a recent work on advanced graph filters that considers the set relationships between eigenvalues, where rich geometric information is embedded.
For example, the algebraic multiplicity of eigenvalue 0 reflects the number of connected components in the graph.
But element-wise graph filters are not aware of such statistics.
Therefore, Specformer uses Transformer \citep{Transformer} as a set-to-set graph filter to learn the set relationships between eigenvalues:
\begin{equation}
    \mathbf{G}=\text{Transformer}(\rho(\bm{\Lambda})),
\end{equation}
where $\rho(\cdot)$ is a positional encoding function to vectorize the scalar eigenvalues.
Theoretically, the set-to-set graph filter is a generalization of element-wise filters and, therefore, more expressive.
However, the quadratic complexity of Transformer further increases the complexity of Specformer and makes it impossible to scale to large graphs.

\noindent \textbf{Summary.}
The advanced filters are the first kind of spectral GNNs, which has great potential in graph representation learning.
However, the explicit matrix factorization is time-consuming and prevents advanced filters from scaling to large graphs. Therefore, it is important to find suitable domains for advanced filters, such as molecular representation learning \citep{molecular}.

\subsubsection{Polynomial Filters}
The polynomial filters are special cases of advanced filters that simplify the neural networks into polynomial functions.
The greatest advantage of the polynomial filters is that they avoid the explicit EVD but still preserve the arbitrary filtering ability.
In addition to reducing the time complexity, ChebNet also learns a localized graph filter, \ie only nodes within $K$-hops are connected.
The localization property ensures that graph filters remain sparsely connected in space and reduce the cost of memory.
Therefore, polynomial filters attract considerable attention in recent years.

The basic polynomial filter is the weighted sum of graph Laplacian of different orders, which can be defined as:
\begin{equation}
\begin{split}
    \mathbf{G} &= \theta_{0}\mathbf{I} + \theta_{1}\bm{\Lambda} + \cdots + \theta_{K}\bm{\Lambda}^{K}, \\
    \mathbf{U}\mathbf{G}\mathbf{U}^{\top} &= \theta_{0}\mathbf{I} + \theta_{1}\mathbf{L} + \cdots + \theta_{K}\mathbf{L}^{K},
\end{split}
\end{equation}
where $\{\theta_{0}, \theta_{1}, \cdots, \theta_{K}\}$ are the coefficients of polynomials.
However, there are some disadvantages. For example, the polynomial terms $\{\mathbf{L}, \mathbf{L}^{2}, \cdots, \mathbf{L}^{K}\}$ are non-orthogonal, which may affect the convergence of models.
There are many works that try to improve the design of polynomial filters, most of which are based on the approximation theory, such as Chebyshev polynomial, Bernstein polynomial, etc.

ChebNet \citep{ChebNet} uses Chebyshev expansion to construct a polynomial filter.
Due to its orthogonality, ChebNet can be trained quickly to converge with a small number of polynomial terms.
The Chebyshev polynomial filter is defined as:
\begin{equation}
    \mathbf{G} = \sum_{k=1}^{K} \theta_{k}\mathcal{T}_{k}(\Tilde{\mathbf{\Lambda}}),
\end{equation}
where $\Tilde{\mathbf{\Lambda}} = \frac{2\mathbf{\Lambda}}{\lambda_{max}} - \mathbf{I}_{N}$ that scales the eigenvalues in the range $[-1, 1]$ and 
$\mathcal{T}_{k}(x)=2x\mathcal{T}_{k-1}(x)-\mathcal{T}_{k-2}(x)$ with $\mathcal{T}_{0}=1, \mathcal{T}_{1}=x$ constructs the orthogonal space.
Although ChebNet has a good approximation ability, it may fail under certain boundary conditions, known as the Runge phenomenon \citep{Runge}.
ChebNetII \citep{ChebNetII} is then proposed to use Chebyshev interpolation to enhance the approximation of ChebNet and alleviate the over-fitting problem.

There are also other polynomial filters. For example, GPR-GNN \citep{GPR-GNN} considers the PageRank information, BernNet \citep{BernNet} uses Bernstein polynomial to keep the graph filter positive semidefinite, JacobiConv \citep{Jacobi} provides a more general orthogonal base through Jacobi polynomial, and DSGC \citep{DSGC} designs polynomial filters on both node and attribute affinity graphs. Here we do not describe them in detail.

Another type of polynomial filter is the relational graph filter, which has a good approximation to the narrow bands, where there is a sharp change in the frequency response.
The aforementioned polynomial filters need to increase their order to fit such narrow bands, which may result in numerical instability.
Relational graph filters solve this problem by learning the ratio between two polynomials as the filtering, which can be formulated as:
\begin{equation}
    \mathbf{G} = \left(\sum_{q=0}^{Q}\theta_{q}\bm{\Lambda}_{i}^{q}\right)
    \left(\mathbf{I} + \sum_{p=0}^{P}\theta_{p}\bm{\Lambda}_{i}^{p} \right)^{-1},
\end{equation}
where $Q$ and $P$ are the orders of the two polynomials.
CayleyNets \citep{CayleyNets} and GraphARMA \citep{GraphARMA} are two traditional relational graph filters in the complex and real domains, respectively.
Relational graph filters can fit the narrow bands with fewer orders. However, the calculation of the inverse of a matrix is time-consuming and unstable, which requires carefully designed constraints, \ie, $1+\sum_{p=0}^{P}\theta_{p}\lambda_{i} \neq 0, \forall i \in N$.

\noindent \textbf{Summary.} Polynomial filters can approximate arbitrary filters, \eg, low-, medium-, and high-pass, and easily scale to large graphs, which are widely used in various applications. For example, multivariate time-series forecasting \citep{TPGNN}.
However, the numerical instability needs to be carefully considered in designing polynomial filters, which is caused by calculating the $K$-th power of the Laplacian matrix. Spec-GN \citep{Spec-GN} provides a spectrum-shrinking method to increase the order of polynomial filters. One can also consider the trigonometric polynomial to avoid the potential numerical problem.

\subsubsection{Linear Filters}
Linear filters further simplify the polynomial filters to yield high scalability and inductive learning ability.
But they lost the capability to approximate arbitrary graph filters.
In general, linear filters can only scale the graph spectrum, but not change the response patterns.

GCN \citep{GCN} is one of the most important works in the field of GNNs, which is the first-order approximation of ChebNet and can be seen as a linear filter. GCN takes the first two terms of ChebNet, \ie, $\theta_{0}\mathbf{I}$ and $\theta_{1}(\mathbf{L}-\mathbf{I})$, and assumes that $\theta = \theta_{0} = -\theta_{1}$. Then it has a linear filter:
\begin{equation}
    \mathbf{U}\mathbf{G}\mathbf{U}^{\top} = \theta(2\mathbf{I} - \mathbf{L})=\theta(\mathbf{I} + \mathbf{A}) \approx \theta \Tilde{\mathbf{A}},
\end{equation}
where $\Tilde{\mathbf{A}}$ is the renormalization trick of adjacency matrix.
Here we can see that the graph filter $\mathbf{G}$ is linear with the input graph structure $\Tilde{\mathbf{A}}$. This explains why GCN has the over-smoothing phenomenon \citep{Deeper}: The parameters of neural networks can only scale the graph spectrum of $\Tilde{\mathbf{A}}$ but do not change its inherent low-pass pattern \citep{SGC}.
By continuously multiplying the graph filter, only the first trivial eigenspace $\mathbf{u}_{1}=[\frac{1}{\sqrt{N}}, \cdots, \frac{1}{\sqrt{N}}]^{\top}$ can preserve the amplitude 1 and others 0.

Although the linear property restricts the expressive power of GCN, it still inspires a series of subsequent works.
PPNP \citep{PPNP} combines GCN with PageRank and adds a residual connection to the graph convolutional layers: 
\begin{equation}
    \mathbf{U}\mathbf{G}\mathbf{U}^{\top}=\theta((1-\alpha)\Tilde{\mathbf{A}} + \alpha \mathbf{I}),
\end{equation}
where $\alpha$ is a hyperparameter to balance the low-pass filter $\Tilde{\mathbf{A}}$ and all-pass filter $\mathbf{I}$.
Compared with GCN, PPNP has a hyperparameter to control the response function of the graph filter.
But PPNP still cannot approximate arbitrary filters.
GNN-LF/HF \citep{GNN-LF} further unifies all linear filters into an optimizing framework.
It decomposes the design of graph convolution into two objectives: fitting and regularization.
By setting different hyperparameters, the fitting term can make the model approximate different filters, \eg, low- and high-pass.
However, such methods can only approximate predefined filters and cannot learn from data, which requires a lot of expert knowledge.

FAGCN \citep{FAGCN} provides a solution to adaptively learn low- or high-pass filters from data but still preserves the linear property.
The key idea is to use neural networks to combine some predefined graph filters.
\begin{equation}
    \mathbf{U}\mathbf{G}\mathbf{U}^{\top} = \theta(\mathbf{I} - \mathbf{M} \odot \mathbf{A}),
\end{equation}
where $\mathbf{M} \in \mathbb{R}^{N \times N}$ is a sparse edge weight matrix learned by neural networks. When the elements in $\mathbf{M}$ is larger than 0, FAGCN can act as a low-pass filter and otherwise a high-pass filter.
There are also some similar methods. For example, AdaGNN \citep{AdaGNN}, AKGNN \citep{AKGNN} and ADA-UGNN \citep{ADA-UGNN}.
Through parameterizing the hyperparameters with neural networks, the linear filters can adaptively approximate different filters. But the expressive power is still not as good as the aforementioned two methods.

\noindent \textbf{Summary.}
Generally, the linear filters are equal to the basic spatial GNNs. Taking GCN as an example, if we assign each feature an independent filter, then the spectral filter becomes:
\begin{equation}
\begin{split}
    \mathbf{h}_{j}^{(l+1)} &= \sigma(\mathbf{U} \sum_{i=1}^{d_{in}}\theta_{ij}\Tilde{\bm{\Lambda}}
    \mathbf{U}^{\top}\mathbf{h}_{i}^{(l)})
    = \sigma(\sum_{i=1}^{d_{in}}\theta_{ij} \mathbf{U} \Tilde{\bm{\Lambda}}
    \mathbf{U}^{\top}\mathbf{h}_{i}^{(l)}) \\
    &= \sigma(\sum_{i=1}^{d_{in}}\theta_{ij} \Tilde{\mathbf{A}}
    \mathbf{h}_{i}^{(l)}) = \sigma(\Tilde{\mathbf{A}}
    \mathbf{h}^{(l)} \bm{\theta}_{j}),
\end{split}
\end{equation}
where $\bm{\theta}_{j} \in \mathbf{R}^{d_{in} \times 1}$ is the parameters of neural networks.
Therefore, linear filters have good scalability, but the fitting ability is heavily restricted.
Linear filters cannot directly capture global or high-order information because the predefined filters are usually designed for first-order information.
It is a promising direction to combine linear filters with complicated filters to learn representations effectively and efficiently.

\subsection{Eigenvector-based Spectral GNNs}
A graph signal can be decomposed into the weighted sum of bases, where the weights indicate the energy in different frequencies.
Therefore, it is important to choose a set of bases that well reflect the distribution of graph signals.
Eigenvector-based spectral GNNs aim to design bases to effectively represent the graph signals in the spectral domain.
In this section, we introduce two important bases: Graph Wavelet and Hermitian Laplacian.
In addition, we also mention the basis encoder, which takes the eigenvectors as input and generates positional encodings for nodes.

\subsubsection{Wavelet Basis}
The Fourier bases, \ie, eigenvectors, are the most commonly used basis function. However, there are two shortcomings of eigenvectors:
First, eigenvectors are dense in space. If graph signals only exist in a subgraph, \eg, diffusion signals, GFT needs to use a large number of bases to approximate such signals.
Second, eigenvectors are non-localized, which cannot reflect the spatial location of the graph signals.
To represent graph signals more efficiently, graph wavelets \citep{wavelets} are proposed, which are sparse and localized in the vertex domain.
The graph wavelets $\bm{\psi}_{s}$ are defined as:
\begin{equation}
    \bm{\psi}_{s} = [\bm{\psi}_{s1}, \cdots, \bm{\psi}_{sN}] = \mathbf{U}\text{diag}(e^{s\lambda_{1}}, \cdots, e^{s\lambda_{N}})\mathbf{U}^{\top},
\end{equation}
where $\bm{\psi}_{si}$ is a basis that represents a diffusion signal centered on node $i$ with the scaling parameter $s$.
Generally, a smaller value $s$ indicates a shorter diffusion, \ie, covering fewer neighbors.
By using different scaling parameters, graph wavelets can construct a multi-scale representation of the graph signals.
Graph wavelets not only reflect the energy distribution in the spectral domain but also indicate its location distribution in the vertex domain, which is a powerful spatial-spectral analysis tool.
There are many spectral GNNs based on graph wavelets.

GWNN \citep{GWNN} proposes a non-parametric spectral GNNs based on the graph wavelet transform (GWT):
\begin{equation}
    \mathbf{h}_{j}^{(l+1)} = \sigma(\bm{\psi}_{s}\sum_{i=1}^{d_{in}}\mathbf{G}_{ij}^{(l)} \bm{\psi}_{s}^{-1}\mathbf{h}_{i}^{(l)}) \quad (j=1 \cdots d_{out}),
\end{equation}
where $\bm{\psi}_{s}^{-1}=\mathbf{U}\text{diag}(e^{-s\lambda_{1}}, \cdots, e^{-s\lambda_{N}})\mathbf{U}^{\top}$.
Since the calculation of $\bm{\psi}_{s}$ and $\bm{\psi}_{s}^{-1}$ are time-consuming, \cite{wavelets} uses Chebyshev polynomials to efficiently approximate the wavelet bases.
\cite{MRA} design orthogonal graph wavelets to provide a multi-resolution framework for graph signal analysis.

\subsubsection{Conjugate Basis}
A basic assumption for spectral GNNs is that the underlying graph structures are symmetric.
The EVD of asymmetric graphs will generate complex eigenvalues and eigenvectors.
However, existing graph filters do not support complex operations.
Therefore, it is important to find a basis that can encode the direction information of edges and have real eigenvalues for graph filtering.
Hermitian Laplacian is a generalization of graph Laplacian, which contains the real part and imaginary parts. The real part is a symmetric matrix, which represents the graph structures. The imaginary part is conjugate, which can be used to encode side information, \eg, edge direction.
The EVD of Hermitian Laplacian has real eigenvalues and complex eigenvectors, which we call conjugate basis.
Therefore, it can be directly incorporated with existing graph filters.

MagNet \citep{MagNet} uses the Hermitian Laplacian to design a spectral GNN for directed graphs.
The normalized Hermitian Laplacian of a directed graph is defined as:
\begin{equation}
    \mathbb{L}=\mathbf{I} - \mathbb{D}^{-\frac{1}{2}}\mathbb{A}\mathbb{D}^{-\frac{1}{2}} \odot \exp\left( \mathrm{i} \bm{\Theta}^{(q)}\right) \quad \mathrm{i} \cdot \mathrm{i} = -1,
\end{equation}
where $\mathbb{A} = \frac{1}{2}(\mathbf{A} + \mathbf{A}^{\top})$, $\mathbb{D}$ is the degree matrix of $\mathbb{A}$ and $\bm{\Theta}^{(q)}=2 \pi q(\mathbf{A} - \mathbf{A}^{\top})$ is the phase matrix, where the incoming and outgoing edges are set to 1 and -1, respectively.
Therefore, they will have opposite values in the complex plane.
The hyparparameter $q$ represents the importance of structural information and direction information. A smaller value of $q$ indicates that the structural information is more important.
Based on the Hermitian Laplacian, MagNet defines a linear filter similar to GCN in the directed graph:
\begin{equation}
    \mathbf{h}_{j}^{(l+1)} = \theta_{ij}\Tilde{\mathbb{A}}\odot \exp\left(\mathrm{i} \bm{\Theta}^{(q)}\right)\mathbf{h}_{i}^{(l)},
\end{equation}
where $\Tilde{\mathbb{A}}$ is the renormalization of $\mathbb{A}$.
MagNet sheds light on how to combine graph filters with the side information of graphs.
It is promising to extend graph filters to other types of graph data by using Hermitian Laplacian. For example, heterogeneous graphs that use different relations as the side information.

\subsubsection{Basis Encoder}
In addition to representing the energy of graph signals, the eigenvectors also reflect the global positions of nodes, which can be used as positional encodings to improve the expressive power of spatial GNNs and break the limitation of 1-WL test \citep{LSPE}.
There are many attempts to learn data representation from the eigenvectors.
For example, spectral clustering \citep{Clustering} aims to use the top-$k$ eigenvectors to represent the manifold of data.
Besides, graph embedding, which aims to embed nodes into a low-dimension space, can be seen as factorizing different graph matrices \citep{NetMF}, where the eigenvectors are used as the node embeddings.
However, such methods are two-stage approaches that cannot update representations through back-propagation.

Basis encoders aim to learn the position representations of nodes from the eigenvectors in an end-to-end manner.
In addition, they also consider the sign and basis ambiguity of eigenvectors \citep{SignNet, PEG} and design models to learn invariant representations.
Although the basic encoders are not necessarily GNNs, there is a close relationship between them.
Spectral GNNs aim to find the important eigenspaces, \ie $\mathbf{u}\mathbf{u}^{\top}$, and basis encoders are designed to find the important eigenvectors that can well describe the node positions.

SAN \citep{SAN} proposes to use Transformer to construct a permutation-equivariant basis encoder. Through the self-attention mechanism, SAN can learn the importance of different eigenvectors, which has a similar idea to spectral GNNs.
SignNet \citep{SignNet} finds that the eigenspaces are invariant to the sign flips and basis symmetries of eigenvectors and uses invariant and equivariant GNNs to learn positional representations from the eigenspaces, which can be seen as a generalization of spectral GNNs.
PEG \citep{PEG} uses the Euclidean distance between eigenvectors to reweight the adjacency matrix, thus avoiding the ambiguity of eigenvectors. PEG encodes the positional information in edges and performs well on the link prediction task.

\noindent \textbf{Summary.}
Compared with eigenvalues-based spectral GNNs, eigenvector-based methods are under-explored, possibly due to theoretical and computational limitations.
A future direction is to combine the basis information with other tasks.
For example, GCN-SVD \citep{GCN-SVD} shows that the low-frequency bases are robust to graph perturbation.

\subsection{Theory}
In this section, we mainly introduce the theoretical progress on spectral GNNs, including expressive power, robustness, and transferability, which may give a deeper understanding of spectral GNNs.

\subsubsection{Expressive power}
The expressive power of MPNNs is proven to be restricted by 1-WL test \citep{GIN}. But the expressive power of spectral GNNs remains under-determined.
\cite{GNNML} proves that spectral graph convolution with nonlinear filtering can break the limit of the 1-WL test and is as powerful as the 3-WL model.
\cite{Jacobi} make a further attempt. They show that spectral filters are universal approximators when satisfying two conditions: no
multiple eigenvalues and no missing frequency components.
Since existing graph filters will map the multiple eigenvalues into the same scalar, this conclusion gives a future direction for designing spectral GNNs.

\subsubsection{Robustness}
It has been proved that GCNs will have different predictions under structural perturbation  \citep{Netattack}, implying that spatial convolutions are vulnerable to attack.
GCN-LFR \citep{GCN-LFR} is the first to verify the vulnerability of graph filters through matrix perturbation theory.
Their theoretical analysis shows that graph filters are more robust against structural perturbations when the eigenvalues fall into a specific interval.
\cite{EvenNet} find that even-order graph filters are more robust than full-order methods and propose EvenNet, which is a graph filter with only even-order terms and performs well on both homophilic and heterophilic datasets.

\subsubsection{Transferability}
Transferability reflects the generalization ability of graph filters across different graphs.
Intuitively, if two graphs have similar spectrums, the graph filters should have representations on both graphs.
\cite{bound1} first proves that spectral filters are transferable on different graphs and can learn similar representations when there is a small perturbation between two graphs, which establishes the connections between transferability and robustness.
\cite{bound2} further proposes a stability bound to measure the stability and transferability of graph filters.

\begin{table*}[t]
\centering
\begin{tabular}{c|c|cccccc}
\toprule
\multicolumn{1}{c}{} & \multicolumn{1}{c}{} & Cora & Citeseer & Pubmed & Chameleon & Squirrel & Actor \\
\midrule
\multirow{3}[2]{*}{Advanced} & SpectralCNN & 83.17±0.87 & 71.06±1.93 & 83.76±0.51 & 55.73±2.66 & 44.79±2.58 & 28.87±3.77\\
& LanczosNet & 88.03±1.70 & 80.55±1.19 & \textbf{90.13±0.49} & 60.96±3.32 & 45.14±1.44 & 36.07±1.38\\
& Specformer & 88.57±1.01 & \textbf{81.49±0.94} & - & \textbf{74.72±1.29} & \textbf{64.64±0.81} & 41.93±1.04\\
\midrule
\multirow{9}[2]{*}{Polynomial} & ChebNet & 86.37±1.71 & 78.65±1.55 & 88.00±0.74 & 66.94±1.59 & 53.15±1.97 & 37.06±2.41\\
& ChebNetII & 88.08±1.17 & 78.46±1.56 & 88.10±0.63 & 71.26±1.28 & \underline{61.92±1.37} & 41.68±1.04\\
& GPRGNN & \textbf{89.61±1.80} & 80.85±1.29 & \underline{90.09±1.15} & 65.27±2.73 & 46.46±2.01 & 39.16±1.39\\
& BernNet & 87.21±1.13 & 79.33±1.59 & 89.00±0.63 & 68.53±3.05 & 51.74±2.37 & 40.72±1.17\\
& JacobiConv & \underline{89.05±1.17} & 80.44±0.95 & 89.51±0.80 & \underline{74.07±1.63} & 57.42±1.94 & 40.82±1.72\\
& Spec-GN & 88.18±0.85 & 80.07±1.48 & 88.51±0.35 & 65.34±3.20 & 50.96±2.19 & 40.49±1.00\\
& DSGC & 88.24±1.53 & 79.75±1.98 & 87.72±0.69 & 51.73±1.95 & 35.68±1.22 & \textbf{42.63±1.32}\\
& ARMA & 87.70±1.25 & 79.02±1.31 & 88.96±0.51 & 60.96±5.66 & 51.10±0.92 & 38.10±2.19\\
& CayleyNet & 86.55±1.61 & 78.40±1.32 &  - & 68.21±1.85 & 51.65±1.98 & \underline{42.26±1.42}\\
\midrule
\multirow{4}[2]{*}{Linear} & GCN & 87.23±0.82 & 79.82±0.74 & 86.43±0.57 & 59.73±1.75 & 46.55±1.02 & 34.01±1.57\\
& FAGCN & 88.03±1.30 & 80.04±0.78 & 89.22±0.59 & 66.52±2.17 & 50.40±1.73 & 39.69±1.57\\
& AdaGNN & 87.32±1.10 & 78.69±1.50 & 88.72±0.47 & 60.85±2.44 & 49.88±1.85 & 37.16±0.93\\
& AKGNN  & 88.28±1.52 & \underline{81.06±0.84} & 88.96±0.50 & 68.53±0.69 & 46.82±0.75 & 35.70±1.20\\
\bottomrule
\end{tabular}%
\caption{Node classification on real-world datasets, where bold and underline indicate the best and runner-up, respectively.}
\label{tab:classify}%
\end{table*}%

\subsection{Application}
In this section, we review the latest applications and explain why they are suitable for spectral GNNs, including multivariate time-series forecasting, neuroscience, and computer vision.

\subsubsection{Multivariate Time-series Forecasting}
Multivariate time-series forecasting task aims to predict multiple correlated time-series simultaneously, which needs to consider both intra-series and inter-series patterns.
Learning a latent graph between different time-series can capture the inter-relations and significantly improve forecasting performance.
StemGNN \citep{StemGNN} uses Discrete Fourier Transform (DFT) and GFT to capture the intra- and inter-patterns in the spectral domain, respectively.
TPGNN \citep{TPGNN} further proposes a temporal polynomial filter to capture the dynamics between different time-series.

\subsubsection{Neuroscience}
Neuroscience studies the nervous system of human brains.
Brain connectivity can be seen as a special brain graph, where nodes are different brain regions, and edges are their connections.
\cite{brain1} introduces how to apply GNNs for network neuroscience, and \cite{brain2} utilizes graph wavelets to construct a multi-scale analysis of brain networks.
In general, neuroscience is a suitable application for spectral GNNs, as it does not require scalability and has a strong need for interpretability.

\subsubsection{Computer Vision}
Many computer vision tasks can improve performance by introducing graph structures, such as detection, classification, recognition, etc. 
\cite{GraphCV} provides a comprehensive summary of vision tasks involving GNNs, and we present two of them that make use of spectral GNNs.

Point clouds can be seen as special 3D graphs, where the nodes are points and edges are their nearest neighbors.
It is natural to generalize the idea of graph filters to 3D graphs.
\cite{GSP_geom} introduces the recent development of GSP in geometric data, and \cite{GSP_attack} reviews the attacks on point clouds from the perspective of the graph spectral domain.
Point cloud tasks, such as segmentation, often need to learn the global structures of nodes. Compared to the layer-by-layer aggregation mechanism of spatial methods, spectral GNNs are better at learning global information through GFT.

Skeleton-based motion prediction is another task aimed at predicting human motions from sensor data, where sensors can be considered as the nodes in a graph, and edges are their spatial connections.
\cite{Skeleton} proposes a graph scattering network to project the sensor signals into different graph spectrum bands to predict human motions.
Through decomposing the sensor signals, spectral GNNs can capture the potential relationship between human motions and basis functions and use it for classification.

\subsection{Benchmark}
We benchmark existing spectral GNNs on six graph datasets, three of which (Cora, Citeseer, Pubmed) are homophilic graphs, while the other three (Chameleon, Squirrel, Actor) are heterophilic graphs.
We implement these spectral GNNs based on our GAMMA-Spec \footnote{https://github.com/liuyang-tian/Spectral-GNN-Library} framework, which provides the implementation and visualization of different graph filters.
We follow the setting of \cite{BernNet} and randomly select 60\% data for training, 20\% for validation and 20\% for testing.
The results are shown in Table \ref{tab:classify}. 

We can see that the advanced filters outperform other filters in four datasets, confirming that the advanced filters have better approximation ability.
However, due to the high complexity, there is less research on this type of filter.
Polynomial filters take the runner-up spot but are more studied than advanced filters, probably because polynomial filters strike a good balance between effectiveness and efficiency.
The performance of linear filters is worse than other filters.
Due to the lack of nonlinear filtering, there is also less research on linear filters.
Therefore, it is an important direction to empower linear filters with adaptive filtering capability.

\section{Conclusion and Future Directions}
In this paper, we introduced the development of spectral GNNs from the perspective of model, theory, and application. Although some progress has been made, spectral GNNs still lag behind spatial GNNs.
To facilitate the development of spectral GNNs, we briefly discuss some promising future directions.

\subsubsection{Information intersection}
As mentioned in the challenges, the spectral information of graphs is informative, such as eigenvalues, eigenvectors, Fourier coefficients, etc.
Therefore, it is important to exploit the connections between different spectral information and fuse them to learn better representations.
Besides, spectral GNNs should not be limited to supervised or semi-supervised learning.
Applying the rich spectral information to unsupervised or self-supervised learning is also a promising direction.
For example, \cite{SpCo} connects graph spectrum and graph data augmentation to improve the performance of graph contrastive learning.

\subsubsection{Learning from signal processing}
Signal processing on Euclidean data, \eg, time series and images, has a long history. While non-Euclidean signal processing is still an emerging field.
Therefore, spectral GNNs can borrow ideas from traditional signal processing methods.
For example, Kalman filtering, Bessel filtering, etc.
In addition, the spectrum of the Euclidean data can be divided into amplitude spectrum and phase spectrum, representing the shape and location information, respectively.
However, in spectral GNNs, most graph spectrums are amplitude spectrums, and the phase spectrums are largely ignored, which motivates the discovery of phase information in spectral GNNs.

\subsubsection{Broader data and applications}
One of the most important advantages of spatial GNN is the flexibility to adapt to different graph data. The message-passing mechanism can be easily extended to heterogeneous graphs and hypergraphs.
However, the idea of signal processing is hard to generalize to other graph data, which hinders the development of spectral GNN to some extent.
Therefore, an important future direction is to extend spectral GNNs to broader graph data and find suitable applications, \eg, biomedical and neuroscience.

\clearpage

\bibliographystyle{named}
\bibliography{ijcai23}

\begin{thebibliography}{}

\bibitem[\protect\citeauthoryear{Balcilar \bgroup \em et al.\egroup
  }{2021}]{GNNML}
Muhammet Balcilar, Pierre H{\'{e}}roux, Benoit Ga{\"{u}}z{\`{e}}re, Pascal
  Vasseur, S{\'{e}}bastien Adam, and Paul Honeine.
\newblock Breaking the limits of message passing graph neural networks.
\newblock In {\em {ICML}}, volume 139, pages 599--608, 2021.

\bibitem[\protect\citeauthoryear{Bessadok \bgroup \em et al.\egroup
  }{2021}]{brain1}
Alaa Bessadok, Mohamed~Ali Mahjoub, and Islem Rekik.
\newblock Graph neural networks in network neuroscience.
\newblock {\em CoRR}, abs/2106.03535, 2021.

\bibitem[\protect\citeauthoryear{Bianchi \bgroup \em et al.\egroup
  }{2022}]{GraphARMA}
Filippo~Maria Bianchi, Daniele Grattarola, Lorenzo Livi, and Cesare Alippi.
\newblock Graph neural networks with convolutional {ARMA} filters.
\newblock {\em {IEEE} TPAMI}, 44(7):3496--3507, 2022.

\bibitem[\protect\citeauthoryear{Bo \bgroup \em et al.\egroup }{2021}]{FAGCN}
Deyu Bo, Xiao Wang, Chuan Shi, and Huawei Shen.
\newblock Beyond low-frequency information in graph convolutional networks.
\newblock In {\em {AAAI}}, pages 3950--3957, 2021.

\bibitem[\protect\citeauthoryear{Bo \bgroup \em et al.\egroup
  }{2023}]{Specformer}
Deyu Bo, Chuan Shi, Lele Wang, and Renjie Liao.
\newblock Specformer: Spectral graph neural networks meet transformers.
\newblock In {\em {ICLR}}, 2023.

\bibitem[\protect\citeauthoryear{Bruna \bgroup \em et al.\egroup
  }{2014}]{SpectralCNN}
Joan Bruna, Wojciech Zaremba, Arthur Szlam, and Yann LeCun.
\newblock Spectral networks and locally connected networks on graphs.
\newblock In {\em {ICLR}}, 2014.

\bibitem[\protect\citeauthoryear{Cao \bgroup \em et al.\egroup
  }{2020}]{StemGNN}
Defu Cao, Yujing Wang, Juanyong Duan, Ce~Zhang, Xia Zhu, Congrui Huang, Yunhai
  Tong, Bixiong Xu, Jing Bai, Jie Tong, and Qi~Zhang.
\newblock Spectral temporal graph neural network for multivariate time-series
  forecasting.
\newblock In {\em NeurIPS}, 2020.

\bibitem[\protect\citeauthoryear{Chang \bgroup \em et al.\egroup
  }{2021}]{GCN-LFR}
Heng Chang, Yu~Rong, Tingyang Xu, Yatao Bian, Shiji Zhou, Xin Wang, Junzhou
  Huang, and Wenwu Zhu.
\newblock Not all low-pass filters are robust in graph convolutional networks.
\newblock In {\em NeurIPS}, 2021.

\bibitem[\protect\citeauthoryear{Chen \bgroup \em et al.\egroup
  }{2022}]{GraphCV}
Chaoqi Chen, Yushuang Wu, Qiyuan Dai, Hong{-}Yu Zhou, Mutian Xu, Sibei Yang,
  Xiaoguang Han, and Yizhou Yu.
\newblock A survey on graph neural networks and graph transformers in computer
  vision: {A} task-oriented perspective.
\newblock {\em CoRR}, abs/2209.13232, 2022.

\bibitem[\protect\citeauthoryear{Chien \bgroup \em et al.\egroup
  }{2021}]{GPR-GNN}
Eli Chien, Jianhao Peng, Pan Li, and Olgica Milenkovic.
\newblock Adaptive universal generalized pagerank graph neural network.
\newblock In {\em {ICLR}}, 2021.

\bibitem[\protect\citeauthoryear{Defferrard \bgroup \em et al.\egroup
  }{2016}]{ChebNet}
Micha{\"{e}}l Defferrard, Xavier Bresson, and Pierre Vandergheynst.
\newblock Convolutional neural networks on graphs with fast localized spectral
  filtering.
\newblock In {\em {NIPS}}, 2016.

\bibitem[\protect\citeauthoryear{Dong \bgroup \em et al.\egroup }{2020}]{GSP1}
Xiaowen Dong, Dorina Thanou, Laura Toni, Michael~M. Bronstein, and Pascal
  Frossard.
\newblock Graph signal processing for machine learning: {A} review and new
  perspectives.
\newblock {\em {IEEE} Signal Process. Mag.}, 37(6):117--127, 2020.

\bibitem[\protect\citeauthoryear{Dong \bgroup \em et al.\egroup
  }{2021}]{AdaGNN}
Yushun Dong, Kaize Ding, Brian Jalaian, Shuiwang Ji, and Jundong Li.
\newblock Adagnn: Graph neural networks with adaptive frequency response
  filter.
\newblock In {\em {CIKM}}, pages 392--401, 2021.

\bibitem[\protect\citeauthoryear{Dwivedi \bgroup \em et al.\egroup
  }{2022}]{LSPE}
Vijay~Prakash Dwivedi, Anh~Tuan Luu, Thomas Laurent, Yoshua Bengio, and Xavier
  Bresson.
\newblock Graph neural networks with learnable structural and positional
  representations.
\newblock In {\em {ICLR}}, 2022.

\bibitem[\protect\citeauthoryear{Entezari \bgroup \em et al.\egroup
  }{2020}]{GCN-SVD}
Negin Entezari, Saba~A. Al{-}Sayouri, Amirali Darvishzadeh, and Evangelos~E.
  Papalexakis.
\newblock All you need is low (rank): Defending against adversarial attacks on
  graphs.
\newblock In {\em {WSDM}}, pages 169--177, 2020.

\bibitem[\protect\citeauthoryear{Epperson}{1987}]{Runge}
James~F Epperson.
\newblock On the runge example.
\newblock {\em The American Mathematical Monthly}, 94(4):329--341, 1987.

\bibitem[\protect\citeauthoryear{Gilmer \bgroup \em et al.\egroup
  }{2017}]{MPNN}
Justin Gilmer, Samuel~S. Schoenholz, Patrick~F. Riley, Oriol Vinyals, and
  George~E. Dahl.
\newblock Neural message passing for quantum chemistry.
\newblock In {\em {ICML}}, volume~70, pages 1263--1272, 2017.

\bibitem[\protect\citeauthoryear{Guo \bgroup \em et al.\egroup
  }{2022}]{molecular}
Zhichun Guo, Bozhao Nan, Yijun Tian, Olaf Wiest, Chuxu Zhang, and Nitesh~V.
  Chawla.
\newblock Graph-based molecular representation learning.
\newblock {\em CoRR}, abs/2207.04869, 2022.

\bibitem[\protect\citeauthoryear{Hamilton \bgroup \em et al.\egroup
  }{2017}]{GraphSAGE}
William~L. Hamilton, Zhitao Ying, and Jure Leskovec.
\newblock Inductive representation learning on large graphs.
\newblock In {\em {NIPS}}, 2017.

\bibitem[\protect\citeauthoryear{Hammond \bgroup \em et al.\egroup
  }{2011}]{wavelets}
David~K Hammond, Pierre Vandergheynst, and R{\'e}mi Gribonval.
\newblock Wavelets on graphs via spectral graph theory.
\newblock {\em Applied and Computational Harmonic Analysis}, 30(2):129--150,
  2011.

\bibitem[\protect\citeauthoryear{He \bgroup \em et al.\egroup }{2021}]{BernNet}
Mingguo He, Zhewei Wei, Zengfeng Huang, and Hongteng Xu.
\newblock Bernnet: Learning arbitrary graph spectral filters via bernstein
  approximation.
\newblock In {\em NeurIPS}, 2021.

\bibitem[\protect\citeauthoryear{He \bgroup \em et al.\egroup
  }{2022}]{ChebNetII}
Mingguo He, Zhewei Wei, and Ji{-}Rong Wen.
\newblock Convolutional neural networks on graphs with chebyshev approximation,
  revisited.
\newblock In {\em NeurIPS}, 2022.

\bibitem[\protect\citeauthoryear{Hu \bgroup \em et al.\egroup
  }{2020}]{GSP_geom}
Wei Hu, Jiahao Pang, Xianming Liu, Dong Tian, Chia{-}Wen Lin, and Anthony
  Vetro.
\newblock Graph signal processing for geometric data and beyond: Theory and
  applications.
\newblock {\em CoRR}, abs/2008.01918, 2020.

\bibitem[\protect\citeauthoryear{Ju \bgroup \em et al.\egroup }{2022}]{AKGNN}
Mingxuan Ju, Shifu Hou, Yujie Fan, Jianan Zhao, Yanfang Ye, and Liang Zhao.
\newblock Adaptive kernel graph neural network.
\newblock In {\em {AAAI}}, pages 7051--7058, 2022.

\bibitem[\protect\citeauthoryear{Kenlay \bgroup \em et al.\egroup
  }{2021}]{bound2}
Henry Kenlay, Dorina Thanou, and Xiaowen Dong.
\newblock Interpretable stability bounds for spectral graph filters.
\newblock In {\em {ICML}}, volume 139, pages 5388--5397, 2021.

\bibitem[\protect\citeauthoryear{Kipf and Welling}{2017}]{GCN}
Thomas~N. Kipf and Max Welling.
\newblock Semi-supervised classification with graph convolutional networks.
\newblock In {\em {ICLR}}, 2017.

\bibitem[\protect\citeauthoryear{Klicpera \bgroup \em et al.\egroup
  }{2019}]{PPNP}
Johannes Klicpera, Aleksandar Bojchevski, and Stephan G{\"{u}}nnemann.
\newblock Predict then propagate: Graph neural networks meet personalized
  pagerank.
\newblock In {\em {ICLR}}, 2019.

\bibitem[\protect\citeauthoryear{Kreuzer \bgroup \em et al.\egroup
  }{2021}]{SAN}
Devin Kreuzer, Dominique Beaini, William~L. Hamilton, Vincent L{\'{e}}tourneau,
  and Prudencio Tossou.
\newblock Rethinking graph transformers with spectral attention.
\newblock In {\em NeurIPS}, 2021.

\bibitem[\protect\citeauthoryear{Lei \bgroup \em et al.\egroup
  }{2022}]{EvenNet}
Runlin Lei, Zhen Wang, Yaliang Li, Bolin Ding, and Zhewei Wei.
\newblock Evennet: Ignoring odd-hop neighbors improves robustness of graph
  neural networks.
\newblock {\em CoRR}, abs/2205.13892, 2022.

\bibitem[\protect\citeauthoryear{Levie \bgroup \em et al.\egroup
  }{2019}]{CayleyNets}
Ron Levie, Federico Monti, Xavier Bresson, and Michael~M. Bronstein.
\newblock Cayleynets: Graph convolutional neural networks with complex rational
  spectral filters.
\newblock {\em {IEEE} TSP}, 67(1):97--109, 2019.

\bibitem[\protect\citeauthoryear{Levie \bgroup \em et al.\egroup
  }{2021}]{bound1}
Ron Levie, Wei Huang, Lorenzo Bucci, Michael~M. Bronstein, and Gitta Kutyniok.
\newblock Transferability of spectral graph convolutional neural networks.
\newblock {\em J. Mach. Learn. Res.}, 22:272:1--272:59, 2021.

\bibitem[\protect\citeauthoryear{Li \bgroup \em et al.\egroup }{2018}]{Deeper}
Qimai Li, Zhichao Han, and Xiao{-}Ming Wu.
\newblock Deeper insights into graph convolutional networks for semi-supervised
  learning.
\newblock In {\em {AAAI}}, pages 3538--3545, 2018.

\bibitem[\protect\citeauthoryear{Li \bgroup \em et al.\egroup
  }{2021a}]{Skeleton}
Maosen Li, Siheng Chen, Zihui Liu, Zijing Zhang, Lingxi Xie, Qi~Tian, and
  Ya~Zhang.
\newblock Skeleton graph scattering networks for 3d skeleton-based human motion
  prediction.
\newblock In {\em {ICCV}}, pages 854--864, 2021.

\bibitem[\protect\citeauthoryear{Li \bgroup \em et al.\egroup }{2021b}]{DSGC}
Qimai Li, Xiaotong Zhang, Han Liu, Quanyu Dai, and Xiao{-}Ming Wu.
\newblock Dimensionwise separable 2-d graph convolution for unsupervised and
  semi-supervised learning on graphs.
\newblock In {\em {KDD}}, pages 953--963, 2021.

\bibitem[\protect\citeauthoryear{Liao \bgroup \em et al.\egroup
  }{2019}]{LanczosNet}
Renjie Liao, Zhizhen Zhao, Raquel Urtasun, and Richard~S. Zemel.
\newblock Lanczosnet: Multi-scale deep graph convolutional networks.
\newblock In {\em {ICLR}}, 2019.

\bibitem[\protect\citeauthoryear{Lim \bgroup \em et al.\egroup
  }{2022}]{SignNet}
Derek Lim, Joshua Robinson, Lingxiao Zhao, Tess~E. Smidt, Suvrit Sra, Haggai
  Maron, and Stefanie Jegelka.
\newblock Sign and basis invariant networks for spectral graph representation
  learning.
\newblock {\em CoRR}, abs/2202.13013, 2022.

\bibitem[\protect\citeauthoryear{Liu \bgroup \em et al.\egroup
  }{2022a}]{GSP_attack}
Daizong Liu, Wei Hu, and Xin Li.
\newblock Point cloud attacks in graph spectral domain: When 3d geometry meets
  graph signal processing.
\newblock {\em CoRR}, abs/2207.13326, 2022.

\bibitem[\protect\citeauthoryear{Liu \bgroup \em et al.\egroup }{2022b}]{SpCo}
Nian Liu, Xiao Wang, Deyu Bo, Chuan Shi, and Jian Pei.
\newblock Revisiting graph contrastive learning from the perspective of graph
  spectrum.
\newblock In {\em NeurIPS}, 2022.

\bibitem[\protect\citeauthoryear{Liu \bgroup \em et al.\egroup }{2022c}]{TPGNN}
Yijing Liu, Qinxian Liu, Jian-Wei Zhang, Haozhe Feng, Zhongwei Wang, Zihan
  Zhou, and Wei Chen.
\newblock Multivariate time-series forecasting with temporal polynomial graph
  neural networks.
\newblock In {\em NeurIPS}, 2022.

\bibitem[\protect\citeauthoryear{Ma \bgroup \em et al.\egroup
  }{2021}]{ADA-UGNN}
Yao Ma, Xiaorui Liu, Tong Zhao, Yozen Liu, Jiliang Tang, and Neil Shah.
\newblock A unified view on graph neural networks as graph signal denoising.
\newblock In {\em {CIKM}}, pages 1202--1211, 2021.

\bibitem[\protect\citeauthoryear{Ng \bgroup \em et al.\egroup
  }{2001}]{Clustering}
Andrew~Y. Ng, Michael~I. Jordan, and Yair Weiss.
\newblock On spectral clustering: Analysis and an algorithm.
\newblock In {\em {NIPS}}, 2001.

\bibitem[\protect\citeauthoryear{Ortega \bgroup \em et al.\egroup
  }{2018}]{GSP2}
Antonio Ortega, Pascal Frossard, Jelena Kovacevic, Jos{\'{e}} M.~F. Moura, and
  Pierre Vandergheynst.
\newblock Graph signal processing: Overview, challenges, and applications.
\newblock {\em Proc. {IEEE}}, 106(5):808--828, 2018.

\bibitem[\protect\citeauthoryear{Qiu \bgroup \em et al.\egroup }{2018}]{NetMF}
Jiezhong Qiu, Yuxiao Dong, Hao Ma, Jian Li, Kuansan Wang, and Jie Tang.
\newblock Network embedding as matrix factorization: Unifying deepwalk, line,
  pte, and node2vec.
\newblock In {\em {WSDM}}, pages 459--467, 2018.

\bibitem[\protect\citeauthoryear{Ramp{\'{a}}sek \bgroup \em et al.\egroup
  }{2022}]{GPS}
Ladislav Ramp{\'{a}}sek, Mikhail Galkin, Vijay~Prakash Dwivedi, Anh~Tuan Luu,
  Guy Wolf, and Dominique Beaini.
\newblock Recipe for a general, powerful, scalable graph transformer.
\newblock {\em CoRR}, abs/2205.12454, 2022.

\bibitem[\protect\citeauthoryear{Spielman}{2007}]{SGT}
Daniel~A. Spielman.
\newblock Spectral graph theory and its applications.
\newblock In {\em {FOCS}}, pages 29--38, 2007.

\bibitem[\protect\citeauthoryear{Vaswani \bgroup \em et al.\egroup
  }{2017}]{Transformer}
Ashish Vaswani, Noam Shazeer, Niki Parmar, Jakob Uszkoreit, Llion Jones,
  Aidan~N. Gomez, Lukasz Kaiser, and Illia Polosukhin.
\newblock Attention is all you need.
\newblock In {\em {NIPS}}, 2017.

\bibitem[\protect\citeauthoryear{Velickovic \bgroup \em et al.\egroup
  }{2018}]{GAT}
Petar Velickovic, Guillem Cucurull, Arantxa Casanova, Adriana Romero, Pietro
  Li{\`{o}}, and Yoshua Bengio.
\newblock Graph attention networks.
\newblock In {\em {ICLR}}, 2018.

\bibitem[\protect\citeauthoryear{Wang and Zhang}{2022}]{Jacobi}
Xiyuan Wang and Muhan Zhang.
\newblock How powerful are spectral graph neural networks.
\newblock In {\em {ICML}}, volume 162, pages 23341--23362, 2022.

\bibitem[\protect\citeauthoryear{Wang \bgroup \em et al.\egroup }{2022}]{PEG}
Haorui Wang, Haoteng Yin, Muhan Zhang, and Pan Li.
\newblock Equivariant and stable positional encoding for more powerful graph
  neural networks.
\newblock In {\em {ICLR}}, 2022.

\bibitem[\protect\citeauthoryear{Wu \bgroup \em et al.\egroup }{2019}]{SGC}
Felix Wu, Amauri H.~Souza Jr., Tianyi Zhang, Christopher Fifty, Tao Yu, and
  Kilian~Q. Weinberger.
\newblock Simplifying graph convolutional networks.
\newblock In {\em {ICML}}, volume~97, pages 6861--6871, 2019.

\bibitem[\protect\citeauthoryear{Wu \bgroup \em et al.\egroup }{2021}]{survey1}
Zonghan Wu, Shirui Pan, Fengwen Chen, Guodong Long, Chengqi Zhang, and
  Philip~S. Yu.
\newblock A comprehensive survey on graph neural networks.
\newblock {\em {IEEE} TNNLS}, 32(1):4--24, 2021.

\bibitem[\protect\citeauthoryear{Xu \bgroup \em et al.\egroup }{2019a}]{GWNN}
Bingbing Xu, Huawei Shen, Qi~Cao, Yunqi Qiu, and Xueqi Cheng.
\newblock Graph wavelet neural network.
\newblock In {\em {ICLR}}, 2019.

\bibitem[\protect\citeauthoryear{Xu \bgroup \em et al.\egroup }{2019b}]{GIN}
Keyulu Xu, Weihua Hu, Jure Leskovec, and Stefanie Jegelka.
\newblock How powerful are graph neural networks?
\newblock In {\em {ICLR}}, 2019.

\bibitem[\protect\citeauthoryear{Xu \bgroup \em et al.\egroup }{2019c}]{brain2}
Wenyan Xu, Qing Li, Zhiyuan Zhu, and Xia Wu.
\newblock A novel graph wavelet model for brain multi-scale
  activational-connectional feature fusion.
\newblock In {\em {MICCAI}}, volume 11766, pages 763--771, 2019.

\bibitem[\protect\citeauthoryear{Yang \bgroup \em et al.\egroup
  }{2022}]{Spec-GN}
Mingqi Yang, Yanming Shen, Rui Li, Heng Qi, Qiang Zhang, and Baocai Yin.
\newblock A new perspective on the effects of spectrum in graph neural
  networks.
\newblock In {\em {ICML}}, volume 162, pages 25261--25279, 2022.

\bibitem[\protect\citeauthoryear{Ying \bgroup \em et al.\egroup
  }{2019}]{GNNExplainer}
Zhitao Ying, Dylan Bourgeois, Jiaxuan You, Marinka Zitnik, and Jure Leskovec.
\newblock Gnnexplainer: Generating explanations for graph neural networks.
\newblock In {\em NeurIPS}, 2019.

\bibitem[\protect\citeauthoryear{Zhang and Chen}{2018}]{SEAL}
Muhan Zhang and Yixin Chen.
\newblock Link prediction based on graph neural networks.
\newblock In {\em NeurIPS}, 2018.

\bibitem[\protect\citeauthoryear{Zhang \bgroup \em et al.\egroup
  }{2021}]{MagNet}
Xitong Zhang, Yixuan He, Nathan Brugnone, Michael Perlmutter, and Matthew~J.
  Hirn.
\newblock Magnet: {A} neural network for directed graphs.
\newblock In {\em NeurIPS}, 2021.

\bibitem[\protect\citeauthoryear{Zheng \bgroup \em et al.\egroup }{2020}]{MRA}
Xuebin Zheng, Bingxin Zhou, Ming Li, Yu~Guang Wang, and Junbin Gao.
\newblock Mathnet: Haar-like wavelet multiresolution-analysis for graph
  representation and learning.
\newblock {\em CoRR}, abs/2007.11202, 2020.

\bibitem[\protect\citeauthoryear{Zhou \bgroup \em et al.\egroup
  }{2020}]{survey2}
Jie Zhou, Ganqu Cui, Shengding Hu, Zhengyan Zhang, Cheng Yang, Zhiyuan Liu,
  Lifeng Wang, Changcheng Li, and Maosong Sun.
\newblock Graph neural networks: {A} review of methods and applications.
\newblock {\em {AI} Open}, 1:57--81, 2020.

\bibitem[\protect\citeauthoryear{Zhu \bgroup \em et al.\egroup }{2021}]{GNN-LF}
Meiqi Zhu, Xiao Wang, Chuan Shi, Houye Ji, and Peng Cui.
\newblock Interpreting and unifying graph neural networks with an optimization
  framework.
\newblock In {\em {WWW}}, pages 1215--1226, 2021.

\bibitem[\protect\citeauthoryear{Z{\"{u}}gner \bgroup \em et al.\egroup
  }{2018}]{Netattack}
Daniel Z{\"{u}}gner, Amir Akbarnejad, and Stephan G{\"{u}}nnemann.
\newblock Adversarial attacks on neural networks for graph data.
\newblock In {\em {KDD}}, pages 2847--2856, 2018.

\end{thebibliography}

\end{document}